\documentclass[journal]{IEEEtran}
\ifCLASSINFOpdf
\else
\fi

\usepackage{graphicx} 

\usepackage{algorithm}
\usepackage{algorithmic}
\usepackage{tikz}
\usepackage{amsmath} 
\usepackage{bbding}
\usepackage{makecell}
\usepackage{xcolor}
\usepackage{xpatch}
\usepackage{colortbl}  
\usepackage{array} 
\usepackage{amssymb}
\usepackage{booktabs, multirow, tabularx}
\usepackage[colorlinks, citecolor=blue, linkcolor=red, urlcolor=magenta]{hyperref}

\definecolor{mygray}{gray}{.9}

\definecolor{clblue}{RGB}{222, 246, 246}
\newcommand{\cc}[1]{\cellcolor{clblue!70}{#1}}

\definecolor{cgray}{RGB}{191, 191, 191}

\newcommand{\rr}[1]{\cellcolor{red!30}{#1}}
\definecolor{clblue}{RGB}{209, 246, 246}

\newcommand{\oo}[1]{\cellcolor{clorange!60}{#1}}
\definecolor{clorange}{RGB}{255, 136, 16}

\definecolor{tabletitle}{gray}{.8}
\definecolor{ours}{gray}{.95}
\definecolor{ggray}{RGB}{127,127,127}

\definecolor{reda}{RGB}{202,0,0}

\definecolor{redb}{RGB}{217,148,143}

\definecolor{myyellow}{RGB}{190,144,0}
\definecolor{mygreen}{RGB}{0,136,51}
\definecolor{myblue}{RGB}{0,102,204}
\newcolumntype{B}{!{\vrule width 1pt}}
\usepackage{pifont}

\begin{document}

%
\title{WeatherCycle: Unpaired Multi-Weather Restoration via Color Space Decoupled Cycle Learning}
%
%
%

\author{Wenxuan Fang, Jiangwei Weng, Jianjun Qian, Jian Yang\textsuperscript{\rm *} and Jun Li\textsuperscript{\rm *}

\thanks{}
\thanks{ }
\thanks{}
\thanks{This work was supported by the National Natural Science Foundation of China under Grant Nos. U24A20330, 62361166670.}
\thanks{Wenxuan Fang, Jiangwei Weng, Jianjun Qian, Jian Yang and Jun Li are with the School of Computer Science and Engineering, Nanjing University of Science and Technology, Nanjing, 210000, China (email: \{wenxuan\_fang, wengjiangwei, csjqian, csjyang, junli\}@njust.edu.cn)}

\thanks{Corresponding authors: Jian Yang and Jun Li.}

\thanks{* indicates Corresponding authors.}

}

\maketitle

\begin{abstract}
Unsupervised image restoration under multi-weather conditions remains a fundamental yet underexplored challenge. While existing methods often rely on task-specific physical priors, their narrow focus limits scalability and generalization to diverse real-world weather scenarios. In this work, we propose \textbf{WeatherCycle}, a unified unpaired framework that reformulates weather restoration as a bidirectional degradation-content translation cycle, guided by degradation-aware curriculum regularization. At its core, WeatherCycle employs a \textit{lumina-chroma decomposition} strategy to decouple degradation from content without modeling complex weather, enabling domain conversion between degraded and clean images. To model diverse and complex degradations, we propose a \textit{Lumina Degradation Guidance Module} (LDGM), which learns luminance degradation priors from a degraded image pool and injects them into clean images via frequency-domain amplitude modulation, enabling controllable and realistic degradation modeling. Additionally, we incorporate a \textit{Difficulty-Aware Contrastive Regularization (DACR)} module that identifies hard samples via a CLIP-based classifier and enforces contrastive alignment between hard samples and restored features to enhance semantic consistency and robustness. Extensive experiments across serve multi-weather datasets, demonstrate that our method achieves state-of-the-art performance among unsupervised approaches, with strong generalization to complex weather degradations.
\end{abstract}

\begin{IEEEkeywords}
Unsupervised image restoration, color space, contrastive leraning.
\end{IEEEkeywords}

%
\IEEEpeerreviewmaketitle

\section{Introduction}

\IEEEPARstart{A}{dverse} weather conditions such as haze, rain, and low-light frequently degrade image quality, posing significant challenges for both autonomous driving perception system and human vision. Early efforts tackled these problems with supervised models tailored to specific degradation types, including image dehazing \cite{sgdn, c2pnet, fang2023multi}, deraining \cite{nerdrain}, and low-light enhancement \cite{mamballie}. While effective under controlled conditions, these task-specific models lack generalization and struggle to cope with real-world scenarios where multiple degradations often co-occur.

\begin{figure}[!t]
        \centering
        \includegraphics[width=0.97\linewidth]{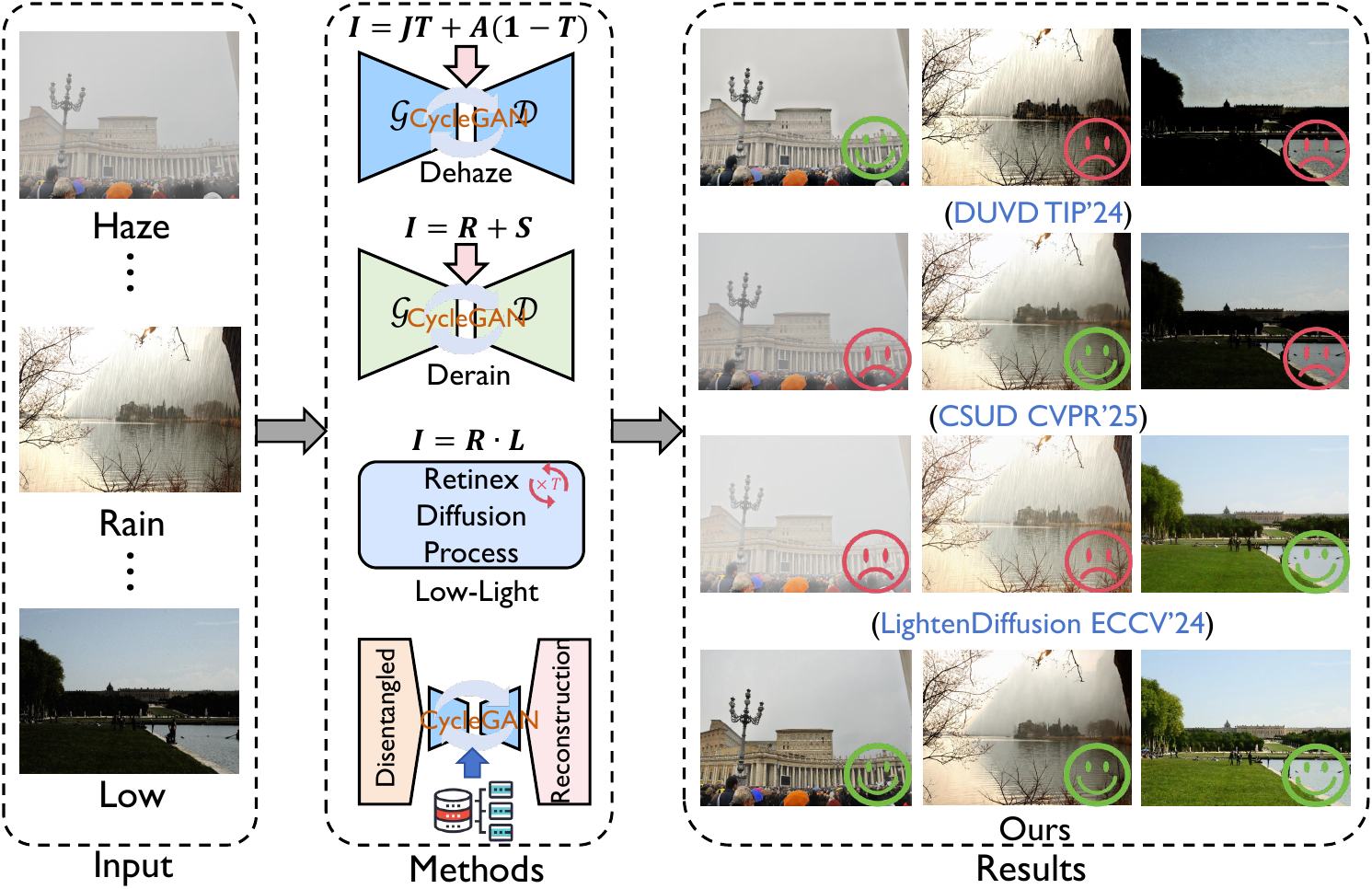}
        \vskip -0.1in 
        \caption{Task-specific unsupervised models (DUVD \cite{duvd}, CSUD \cite{csud}, LightenDiffusion \cite{lightendiffusion}) are limited to single-weather restoration and often fail under unseen degradations. In contrast, our unified framework generalizes across haze, rain, and low-light conditions, producing consistently superior results with better detail preservation and visual quality.}
        \label{fig:CompareMethods}
\end{figure}

Recent research has introduced All-in-One image restoration frameworks that aim to handle multiple weather degradations within a unified model, such as task-specific head \cite{transweather}, weather condition queries \cite{promptir, chen2022learning}, or language-based prompts \cite{guo2025onerestore, daclip} to guide restoration. Despite promising results, such approaches are heavily dependent on large-scale paired datasets with pixel-level alignment across various weather types. However, capturing perfectly aligned pairs in outdoor scenes is expensive and often infeasible. To mitigate this data dependency, several studies have adopted unsupervised image-to-image translation CycleGAN \cite{cyclegan} framework, which use adversarial and cycle-consistency losses to map degraded images to the clean domain. These include approaches for dehazing via atmospheric modeling \cite{duvd}, rain removal through reflection decomposition \cite{csud}, and low-light enhancement with Retinex priors \cite{lightendiffusion}. However, these methods are typically limited to a single degradation type and fail to generalize across weather variations. As shown in Figure~\ref{fig:CompareMethods}, such models remain tightly coupled to specific physical assumptions and lack the flexibility to handle compound or unseen degradations. To the best of our knowledge, there is still no prior work that explicitly addresses the unsupervised all-in-one image restoration problem across diverse weather degradations.

\begin{figure*}[!t]
        \centering
        \includegraphics[width=0.97\linewidth]{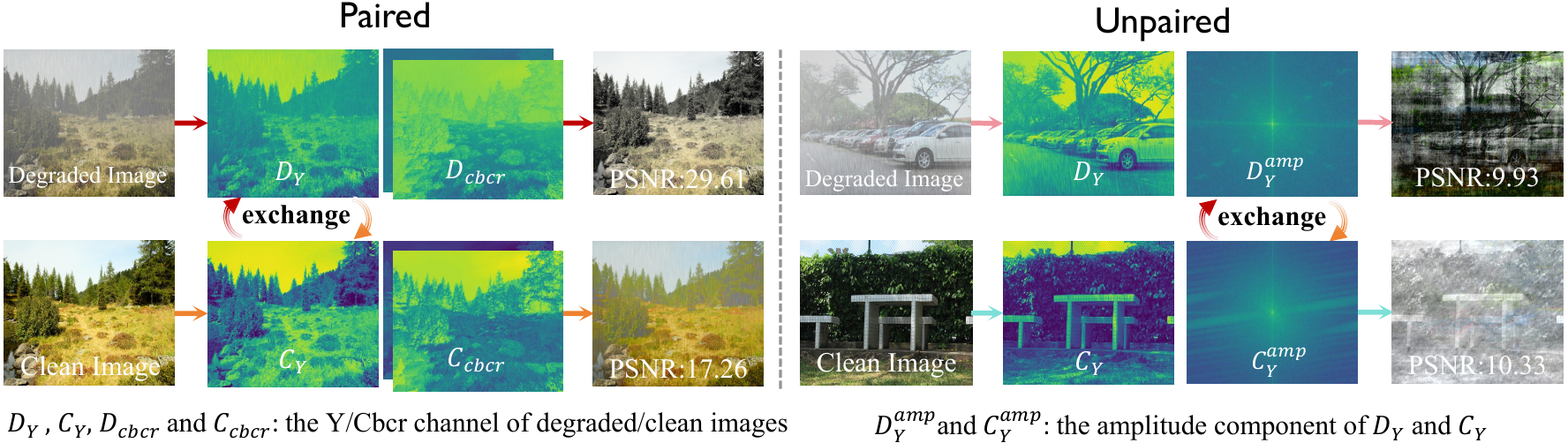}
        \vskip -0.1in 
        \caption{Analysis of degradation concentration in the Y channel. In the paired setting (left), exchanging the Y and CbCr channels between degraded and clean images achieves effective restoration with high PSNR. In the unpaired setting (right), to introduce degradation information correctly, we exchange the amplitude components of the Y channel in the frequency domain, which helps transfer degradation priors.}
        \label{fig:Motivation}
\end{figure*}

In this work, we conduct an in-depth analysis of image statistics under various adverse weather conditions and uncover a consistent trend: most visual degradation is concentrated in the luminance (Y) channel, while chrominance (Cb/Cr) channels remain relatively stable. As illustrated in Figure \ref{fig:Motivation} (left), in the paired setting, directly swapping the Y channels between degraded and clean images significantly improves perceptual quality. We further extend this insight to the more challenging unpaired setting by introducing a frequency-domain strategy: exchanging the amplitude of the Y channel allows effective transfer of degradation priors while preserving structural information (Figure \ref{fig:Motivation}, right).


Based on this key observation, we propose the first unified framework for unpaired multi-weather image restoration with the Lumina-Chroma Decomposition, named WeatherCycle. Our method introduces two symmetric cyclic paths that map between degraded and clean domains. Each pathway performs Y/CbCr decomposition, enabling structure-preserving bidirectional mapping. To enhance degradation modeling under unpaired conditions, we propose a Lumina Degradation Guidance Module (LDGM). It extracts degradation-aware priors from unpaired degraded pool and modulates the luminance amplitude in the frequency domain, guiding structure-preserving domain translation under diverse weather conditions. In addition, we design a Difficulty-Aware Contrastive Regularization (DACR) module, which identifies hard samples via a CLIP-based classifier and enforces contrastive alignment between restored and degraded features, enhancing semantic consistency and robustness. Extensive experiments on multiple benchmark datasets demonstrate that our method outperforms both task-specific baselines and recent all-in-one unsupervised approaches in terms of restoration quality and generalization ability. Our main contributions are summarized as follows:

\begin{itemize}
    \item We propose the first unified framework for multi-weather image restoration with Y/CbCr decomposition under unpaired settings, named WeatherCycle.

    \item A Limina Degradation Guidance Module (LDGM) is designed that adaptively models degradation by injecting frequency-domain amplitude cues from randomly sampled degraded images. This enables flexible and accurate prior transfer, enhancing restoration performance under unpaired settings.

    \item We present a Difficulty-Aware Contrastive Regularization (DACR) that improves restoration under complex degradations by enforcing semantic–geometric consistency through a difficulty classifier and difficulty-adaptive contrastive loss.

    \item Our method achieves superior restoration quality and generalization compared to both task-specific and all-in-one baselines on diverse benchmarks.
\end{itemize}

\begin{figure*}[!t]
        \centering
        \includegraphics[width=0.97\linewidth]{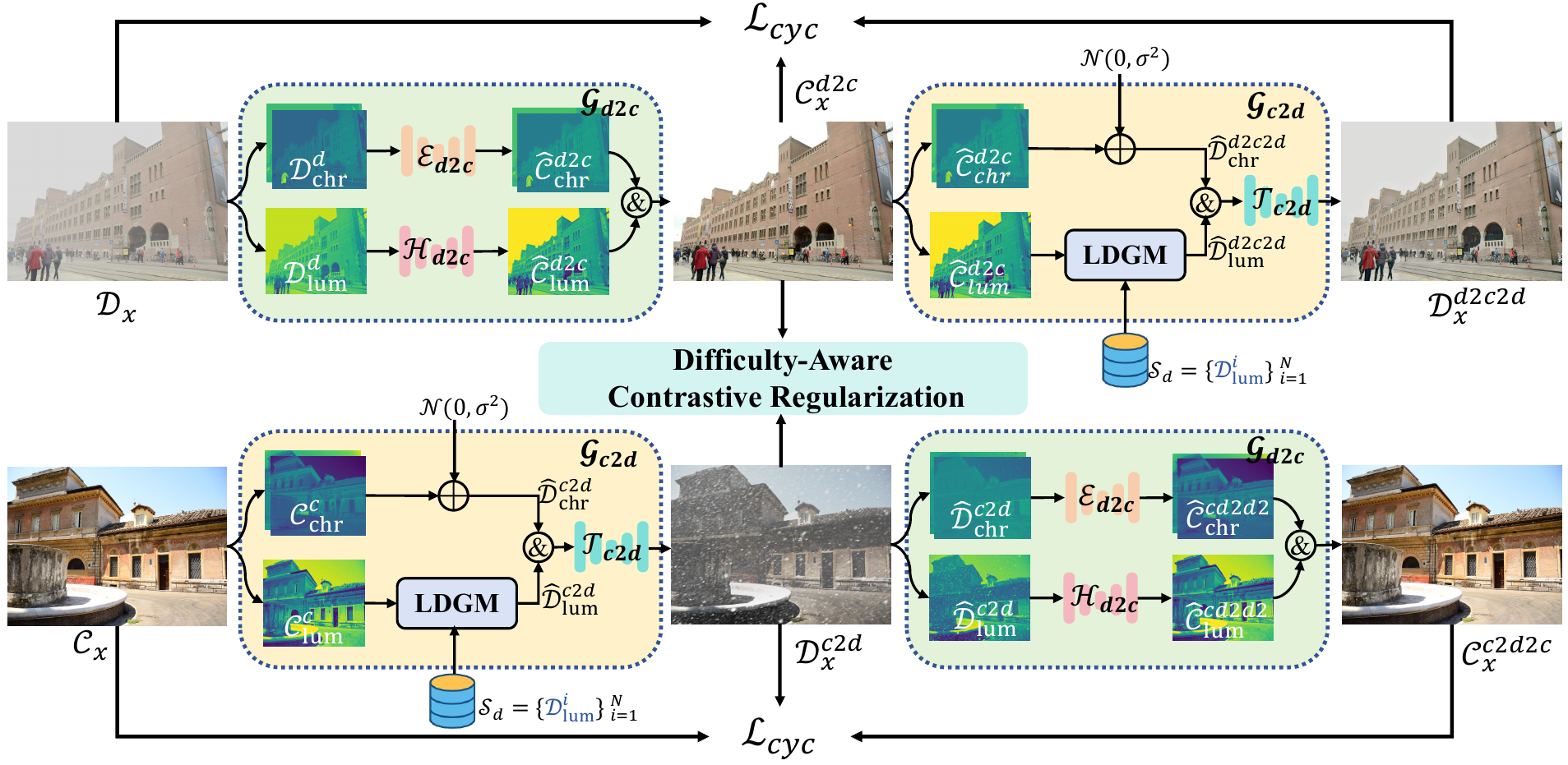}
        \vspace{-0.2cm}
        \caption{Overall architecture of the proposed WeatherCycle. The model consists of two symmetric cycles: a restoration cycle $\mathcal{G}_{d2c}$ that maps degraded images $\mathcal{D}_x$ to clean images $\mathcal{C}_x$, and a re-degradation cycle $\mathcal{G}_{c2d}$ that reconstructs degradation from clean images. Each branch operates on decomposed luminance and chrominance components in the YCbCr space. $\&$ denotes concatenation and inverse YCbCr transformation.}
        \label{fig:method}
\end{figure*}

\section{Related Works}

\subsection{Image restoration in adverse weather conditions.}
Unified image restoration targets various degradations (e.g., rain, snow, haze) within a single model. Early supervised works explored NAS-based multi-task learning~\cite{nas, allinone}, followed by advances such as knowledge distillation~\cite{chen2022learning}, prompt learning~\cite{promptir, daclip, guo2025onerestore}, and conditional diffusion~\cite{t3diffweather, chen2025unirestore, kong2025dual, li2025diffusion}. To reduce dependency on paired data, unsupervised image-to-image translation techniques such as CycleGAN~\cite{cyclegan} have been adopted for weather-specific restoration tasks, including unpaired image dehazing \cite{d4, duvd, wang2025dehaze}, image deraining \cite{deraincyclegan}, and low-light enhancement \cite{lightendiffusion, enlightengan, nerco, weng2025daytime}. For example, Luo et al. proposed WSRRGAN \cite{luo2020weakly}, which utilized a pre-trained VGG classifier to capture the differences between raindrop and clean images, and generates a pseudo-attention map for raindrop images through category activation mapping. Dehaze-RetinexGAN \cite{wang2025dehaze} converted the image dehazing task into a light-reflectance decomposition task based on the dual correlation between Retinex and the dehazing task. Nevertheless, these approaches are tailored to single weather types and lack the flexibility to handle multiple degradations within a unified framework. To the best of our knowledge, no existing work has explicitly addressed multi-weather image restoration under an unpaired setting. Our work fills this gap by introducing a unified unpaired framework that leverages lumina-chroma decomposition and difficulty-aware contrastive learning. Without the need for explicit modeling of each degradation type, our method enables robust and generalizable restoration across diverse weather-induced degradations.

\subsection{Contrastive Learning.}
Contrastive learning has emerged as a powerful paradigm in representation learning by pulling semantically similar pairs closer and pushing dissimilar pairs apart in the feature space. It has been widely applied in unsupervised visual pretraining~\cite{liang2025autoregressive}, cross-domain adaptation \cite{fan2025controlled}, and image deraining \cite{nlcl}. In low-level vision, recent works have introduced contrastive objectives to enforce structure or content consistency across restored and degraded samples. For example, frequency-aware contrastive learning \cite{gao2024efficient} and spatial patch \cite{odcr} contrast have shown promising results in deraining and dehazing tasks. However, these methods typically treat all samples equally, ignoring variations in degradation severity. In contrast, we propose a Difficulty-Aware Contrastive Regularization that adaptively emphasizes harder cases, encouraging the model to learn more robust and generalizable representations under varying weather conditions.

\section{Methods}


In this section, we introduce WeatherCycle, a lumina-chroma decomposition-based cycle-consistent generative adversarial network with degradation-aware guidance. Given a clean image $\mathcal{C}_x$ from the clean domain $\mathcal{X}_C$ and a degraded image $\mathcal{D}_x$ from the degradation domain $\mathcal{X}_D$, WeatherCycle performs two cyclic mappings: 
$\mathcal{C}_x \rightarrow \mathcal{D}_x^{c2d} \rightarrow \mathcal{C}_x^{c2d2c}$ and 
$\mathcal{D}_x \rightarrow \mathcal{C}_x^{d2c} \rightarrow \mathcal{D}_x^{d2c2d}$,
enabling the learning of a restoration network $\mathcal{G}_{d2c}$ that maps degraded inputs to clean outputs without requiring paired supervision.

As illustrated in Figure~\ref{fig:method}, the proposed framework consists of two collaborative modules: the \textbf{Restoration Module} and the \textbf{Re-degradation Module}. 
The restoration module $\mathcal{G}_{d2c}$ decomposes the input into luminance and chrominance components and performs targeted recovery for each channel. 
To model degradation in reverse, the re-degradation module $\mathcal{G}_{c2d}$ learns to transform clean images back into their degraded counterparts, ensuring domain completeness and facilitating cycle-consistent training. To inject realistic luminance degradation without paired data, we introduce a Limina Degradation Guidance Module (LDGM), which samples degradation priors and injects them into the luminance stream guided by frequency-domain amplitude cues. Finally, we design Difficulty-Aware Contrastive Regularization (DACR), which enforces geometric alignment by pulling degraded and restored features closer in feature space.

\subsection{Restoration Module}
The design of our restoration branch is motivated by a key empirical observation from Figure~\ref{fig:Motivation}: image degradation caused by adverse weather conditions is primarily concentrated in the luminance channel, while the chrominance components remain relatively unaffected. This property suggests that focusing the restoration process on the luminance channel enables the model to effectively correct structural distortions, contrast loss, and weather-induced blur, while maintaining color fidelity through minimal adjustment of the chrominance channels. Furthermore, this decomposition alleviates the complexity of modeling diverse weather degradations under unpaired supervision by decoupling structural restoration and color preservation.

Based on this insight, each degraded images $\mathcal{D}_x \in \mathbb{R}^{3\times H \times W}$ is transformed into the YCbCr color space and decomposed into a luminance map $\mathcal{D}^{d}_{\text{lum}} \in \mathbb{R}^{1\times H \times W}$ and a chrominance map $\mathcal{D}^{d}_{\text{chr}} \in \mathbb{R}^{2\times H \times W}$. The luminance stream is processed by a structure-aware backbone $\mathcal{H}_{d2c}$, implemented using NAFNet~\cite{nafnet}, to recover fine-scale structures and contrast. The chrominance stream is refined using a lightweight encoder-decoder $\mathcal{E}_{d2c}$ to preserve semantic color consistency. The restored components $\mathcal{\hat{C}}^{d2c}_{\text{lum}}$ and $\mathcal{\hat{C}}^{d2c}_{\text{chr}}$ are then fused via inverse YCbCr transformation to reconstruct the clean image:
\begin{equation}
    \begin{split}
    & \mathcal{\hat{C}}^{d2c}_{\text{lum}}, \mathcal{\hat{C}}^{d2c}_{\text{chr}} = \mathcal{H}_{d2c}(\mathcal{D}^{d}_{\text{lum}}),  \mathcal{E}_{d2c}(\mathcal{D}^{d}_{\text{chr}}), \\
    & \mathcal{C}^{d2c}_x = \&(\mathcal{\hat{C}}^{d2c}_{\text{lum}}, \mathcal{\hat{C}}^{d2c}_{\text{chr}}).
    \end{split}
\end{equation}
where \& is concat and inverse YCbCr transformation.

\subsection{Re-degradation Module}

To enable unpaired training and ensure domain completeness, we design a re-degradation module $\mathcal{G}_{c2d}$ to simulate realistic degradations from clean inputs. Given a clean image $\mathcal{C}_x \in \mathbb{R}^{3\times H \times W}$, we first convert it to the YCbCr color space and decompose it into a luminance component $\mathcal{C}^{c}_{\text{lum}}$ and a chrominance component $\mathcal{C}^{c}_{\text{chr}}$. To consider for mild degradation in the chrominance channel, we inject additional Gaussian noise $n \sim \mathcal{N}(0, \sigma^2)$ to simulate realistic color degradation, yielding $\hat{\mathcal{D}}^{c2d}_{\text{chr}}$. In contrast, the luminance component is passed through the proposed LDGM to generate a guided degraded map $\hat{\mathcal{D}}^{c2d}_{\text{lum}}$. Finally, the degraded luminance and chrominance components are fused via a learnable reconstruction network $\mathcal{J}^{c2d}$ to restore the final degraded image:
\begin{equation}
    \begin{split}
        & \hat{\mathcal{D}}^{c2d}_{\text{chr}}, \hat{\mathcal{D}}^{c2d}_{\text{lum}} = \mathcal{C}^{c}_{\text{chr}} + n, \mathcal{LD}(\mathcal{C}^{c}_{\text{lum}}), \\
        & D_{x}^{c2d} = \mathcal{J}_{c2d}(\&(\hat{\mathcal{D}}^{c2d}_{\text{chr}}, \hat{\mathcal{D}}^{c2d}_{\text{lum}})
    \end{split}
\end{equation}

\begin{figure}[!t]
        \centering
        \includegraphics[width=0.97\linewidth]{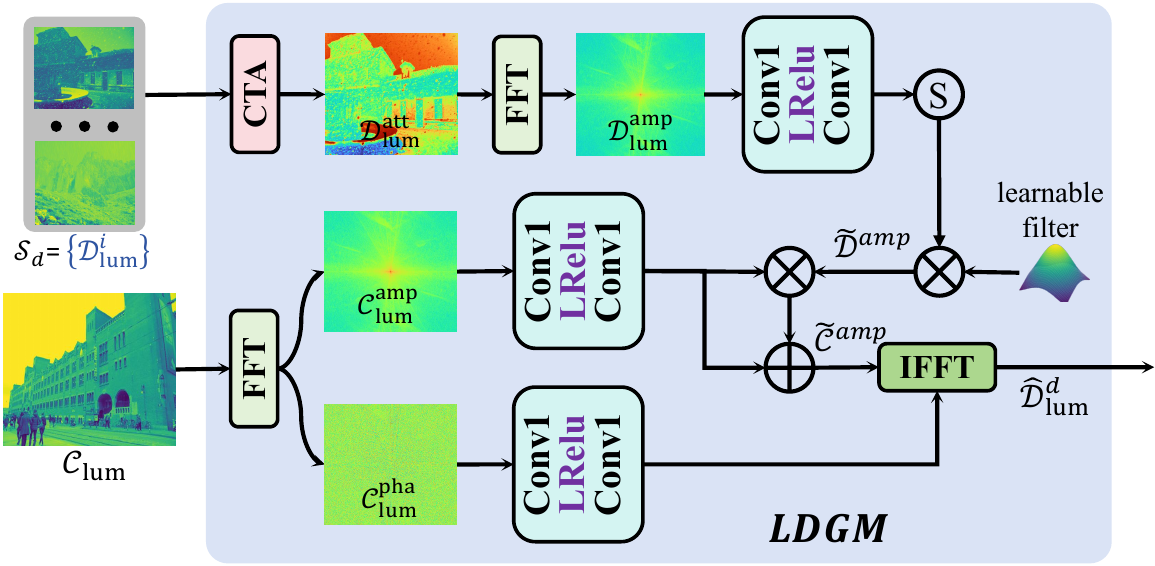}
        \vskip -0.1in 
        \caption{The details of the Limina Degradation Guidance Module (LDGM). CTA is the Channel TOPK Attenion \cite{chen2023learning}.}
        \label{fig:LDGM}
\end{figure}

\textbf{Limina Degradation Guidance Module.} In unpaired image restoration, enforcing consistent degradation modeling is challenging due to the absence of aligned supervision. As show in Figure \ref{fig:Motivation}, we propose to exchange the amplitude components of the Y channel in the frequency domain. This preserves structural priors while enabling effective degradation information transfer. Motivated by this insight, we design the Limina Degradation Guidance Module (LDGM) to inject frequency-guided degradation priors into the clean-to-degraded branch. As shown in Figure~\ref{fig:LDGM}, the module takes as input a clean luminance map $\mathcal{C}_{\text{lum}}$ and sampled degraded patch $\mathcal{D}_{\text{lum}}^{i}$ from a degradation pool $\mathcal{S}_d=\{ \mathcal{D}_{\text{lum}}^{i}\}_{i=1}^{N}$.

The degraded sample $\mathcal{D}_{\text{lum}}^{i}$ is first processed by a channel TOPK attention 
(CTA) mechanism \cite{chen2023learning} to emphasize salient frequency components $\mathcal{D}^{\text{att}}_{\text{lum}}$, which is then transformed into the frequency domain via FFT to obtain the amplitude spectrum $\mathcal{D}^{\text{amp}}_{\text{lum}}$. Meanwhile, the clean luminance input is decomposed into amplitude $\mathcal{C}^{\text{amp}}_{\text{lum}}$ and phase $\mathcal{C}^{\text{pha}}_{\text{lum}}$ components. Each component is separately encoded using convolutional layers with LeakyReLU activations. The degraded amplitude is then passed through a sigmoid activation and further modulated by a learnable spatial filter $\gamma$. The filtered result is element-wise fused with the clean amplitude representation using both addition and multiplication operations.
\begin{equation}
    \begin{split}
        \tilde{\mathcal{D}}^{\text{amp}} &= \sigma(C_1(LR(C_1(\mathcal{D}^{\text{amp}}_{\text{lum}})))) \otimes \gamma, \\
\hat{\mathcal{C}}^{\text{amp}} &= \mathcal{C}^{\text{amp}}_{\text{lum}} \otimes \tilde{\mathcal{D}}^{\text{amp}} + \mathcal{C}^{\text{amp}}_{\text{lum}},
    \end{split}
\end{equation}
where $\tilde{\mathcal{D}}^{\text{amp}}$ is the filtered degradation prior. $\hat{\mathcal{C}}^{\text{amp}}$ is the frequency-guided modulated amplitude. $C_1$ and $LR$ are the $1\times1$ convolutional and LeakyReLU operation, respectively. $\sigma(\cdot)$ is the sigmoid activation function. $\otimes$ denotes element-wise multiplication.

Finally, the fused frequency representations are transformed back into the spatial domain via an inverse FFT (IFFT), resulting in a guided luminance degradation map $\hat{\mathcal{D}}^{d}_{\text{lum}}$, which is subsequently injected into the reconstruction network $\mathcal{J}^{c2d}$.

\subsection{Difficulty-Aware Contrastive Regularization}

While unsupervised models typically rely on adversarial discriminators to distinguish distributions, they often suffer from training instability. Moreover, cycle consistency loss alone is insufficient to enforce semantic and geometric alignment between restored and degraded images. To address these limitations, we propose Difficulty-Aware Contrastive Regularization (DACR), which imposes additional constraints on hard examples in a pretrained semantic feature space to improve restoration quality and generalization.

\begin{figure}[!t]
        \centering
        \includegraphics[width=0.97\linewidth]{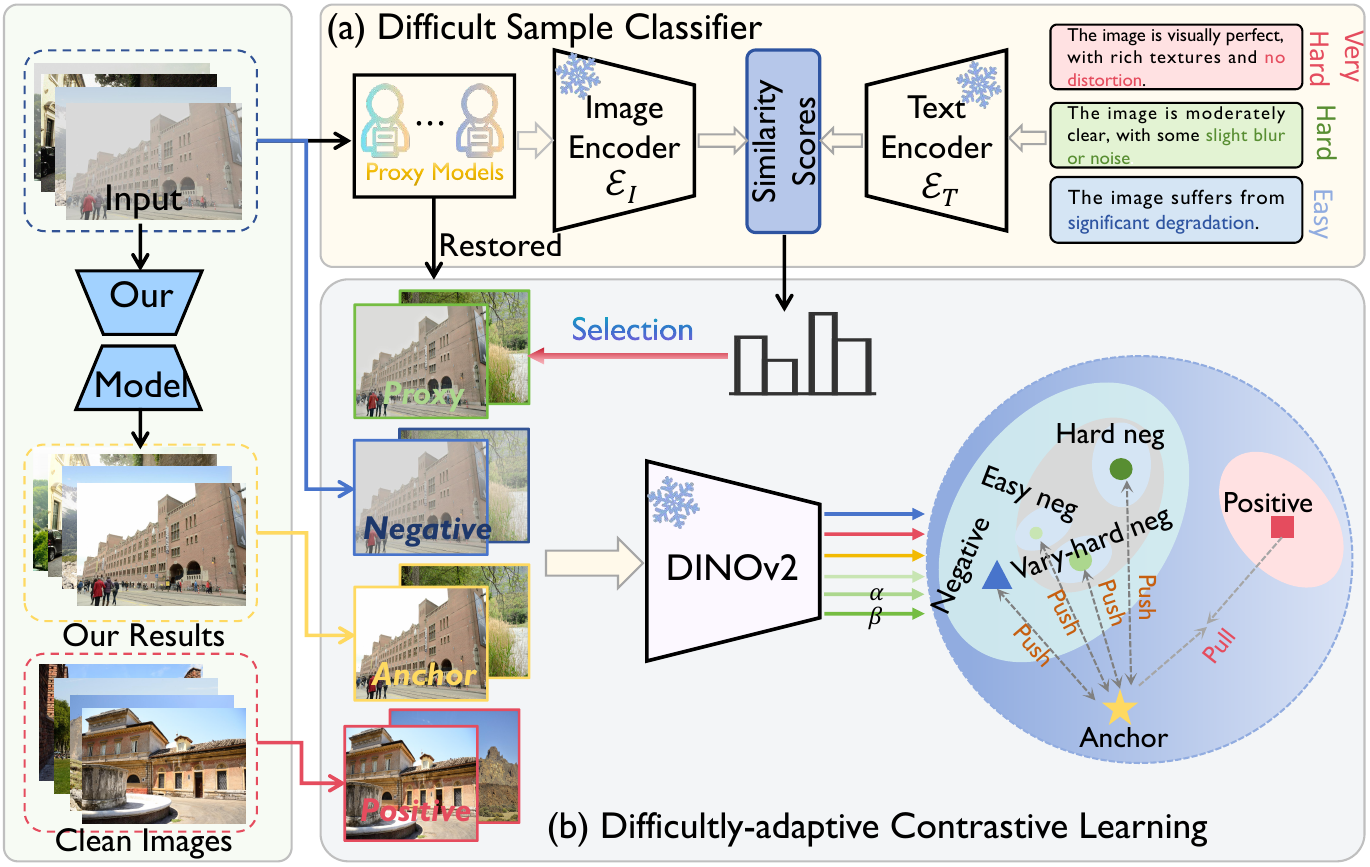}
        \vskip -0.1in 
        \caption{Overview of the proposed Difficulty-Aware Contrastive Regularization (DACR). (a) The difficulty classifier categorizes restoration outputs from proxy models by computing similarity scores between visual embeddings (from $\mathcal{E}_I$) and predefined textual embeddings (from $\mathcal{E}_T$). (b) Contrastive training is performed in the depth feature space, where the restored image serves as the anchor, clean unpaired images as positives, and degraded inputs as negatives. Harder negatives are emphasized using difficulty-aware weights.}
        \label{fig:DACR}
\end{figure}

Our method is based on two key insights: (1) hard cases should be explicitly identified and regularized to focus learning capacity; (2) clearly distinguish degraded images from sharp images in the latent space. We leverage the DINOV2 to enforce structure-aware contrastive alignment. As shown in Figure~\ref{fig:DACR}, we define a set of textual prompts that represent varying degradation levels. First, initial restoration results are generated using proxy models \cite{promptir, fsnet}. These outputs are then processed by a pretrained image encoder $\mathcal{E}_I$ and compared with prompt embeddings from a text encoder $\mathcal{E}_T$ using cosine similarity. The prompt with the highest score determines the difficulty level (e.g., easy-neg, hard-neg, or very-hard).

\begin{table*}[!t]
    \centering
       \caption{Multiple weather restoration comparisons, including image desnowing, deraining, deraining \& dehazing. The 1st and 2nd best results are emphasized with \textcolor{red!50}{\textbf{red}} and \textcolor{clorange}{orange} color, respectively.}
       \vskip -0.1in 
    \setlength{\tabcolsep}{5.5pt}
    \renewcommand{\arraystretch}{1.1}
    \scalebox{0.96}{
    \begin{tabular}{p{0.3cm}|c|c|cc|cc|cc|cc|cc}
 \Xhline{1.2pt} 
           \multirow{2}{*}{\rotatebox[origin=c]{90}{Type}} & \multirow{2}{*}{Methods} 
          & \multirow{2}{*}{Venue} 
          & \multicolumn{2}{c|}{Snow100K-S} 
          & \multicolumn{2}{c|}{Snow100K-L}
          & \multicolumn{2}{c|}{RainDrop} 
          & \multicolumn{2}{c|}{Outdoor-Rain}
          & \multicolumn{2}{c}{\textbf{Avg}} \\ 
        
          ~ & ~  & ~ & PSNR & SSIM & PSNR & SSIM & PSNR & SSIM & PSNR & SSIM & PSNR & SSIM  \\ \hline \hline

         \multirow{6}{*}{\rotatebox[origin=c]{90}{Paired}} & WGWS-Net \cite{zhu2023learning} & CVPR'23 & 36.11 & 0.949 & 29.71 & 0.890 & 31.31 & 0.930 & 25.31 & 0.900 & 30.61 & 0.917\\


        ~ & WeatherDiff$_{64}$ \cite{weatherdiff}  & PAMI'23 & 35.83 & 0.956 & 30.09 & 0.904 & 30.71 & \oo{\textbf{0.931}} & 29.64 & 0.931& 31.56 & 0.930\\ 
        
        ~ & PromptIR \cite{promptir}  & NIPS'23  & 34.60 & \oo{\textbf{0.956}}  & 30.52 & 0.826 & 30.70 & 0.913 & \textbf{\oo{30.95}} & 0.922  & 31.69 & 0.904\\


         ~ & OneRestore \cite{guo2025onerestore}  & ECCV'24 & \oo{\textbf{36.80}} & \oo{\textbf{0.956}} & \oo{\textbf{31.46}} & \oo{\textbf{0.906}} & \oo{\textbf{31.77}} & \oo{\textbf{0.931}} & 30.49& \textbf{\oo{0.932}} & \oo{\textbf{32.63}} & \textbf{\oo{0.931}} \\
            
          ~ &T$^{3}$-DiffWeather \cite{t3diffweather}  & ECCV'24 & \textbf{\rr{37.51}} & \textbf{\rr{0.966}}  & \textbf{\rr{32.37}}  & \textbf{\rr{0.935}}  &  \textbf{\rr{32.66}}  & \textbf{\rr{0.941}} & \textbf{\rr{31.99}}  & \textbf{\rr{0.936}}  & \textbf{\rr{33.63}} & \textbf{\rr{0.944}} \\
          
           ~ & TransWeather-TUR \cite{wu2025debiased} & AAAI'25 & 34.14 & 0.937 & 30.32 & 0.892 & 31.61 & 0.933 & 29.75 & 0.907 & 30.67 & 0.907 \\

          \hline

          \multirow{9}{*}{\rotatebox[origin=c]{90}{Unpaired}} & CycleGAN \cite{cyclegan}  & ICCV'17 & 23.12 & 0.801  & 20.16 & 0.686 & 24.82 & 0.860 &  18.94 & 0.898  & 21.76 & 0.811 \\ 

           ~ & CUT \cite{cut} & ECCV'20 & 23.51 & 0.801 & 20.65 & 0.714 & 23.94 &0.810 & 19.05& 0.688 & 21.78 & 0.753 \\

           ~ & DCLGAN \cite{dclgan} & CVPR'21 & 22.73 & 0.783 & 20.81 & 0.682 & 23.92 &0.818& 19.30 & 0.657 & 21.69 & 0.735 \\

          ~ & Decent \cite{decent}  & NIPS'22 & 24.82 & 0.820 & 21.55 & 0.728 & 24.86 &0.846 & 21.57 & 0.729 & 23.20 & 0.780 \\

          ~ & Santa \cite{santa} & CVPR'23 & 24.10 & 0.759 &  21.48& 0.683 & 23.78 &0.782 & 19.13 & 0.663 & 22.12 &  0.721 \\

        ~ & RCOT \cite{rcot}  & ICML'24 & 33.44 & 0.931 & 26.14 & 0.871 & 28.11 & 0.891 & 26.94 & 0.893 & 28.65 & 0.896 \\

        ~ & DA-RCOT \cite{darcot} & TPAMI'25 & \textcolor{clorange}{33.91} & \textcolor{clorange}{0.936}  & \textcolor{clorange}{27.03} & \textcolor{clorange}{0.886} & \textcolor{clorange}{28.73} & \textcolor{clorange}{0.897} & \textcolor{clorange}{27.72} &  \textcolor{clorange}{0.914}& \textcolor{clorange}{29.34}  & \textcolor{clorange}{0.908} \\ \hline
          
          ~ & \textbf{WeatherCycle (Ours)}  & - & \textcolor{red!50}{\textbf{34.63}} &\textcolor{red!50}{\textbf{0.940}}  & \textcolor{red!50}{\textbf{28.64}}  & \textcolor{red!50}{\textbf{0.898}}  & \textcolor{red!50}{\textbf{29.56}}  & \textcolor{red!50}{\textbf{0.900}}  & \textcolor{red!50}{\textbf{28.46}}  & \textcolor{red!50}{\textbf{0.929}} & \textcolor{red!50}{\textbf{30.19}} & \textcolor{red!50}{\textbf{0.916}} \\
          \Xhline{1.2pt} 
    \end{tabular}
  }
    \label{table:all_weather_result}
\end{table*}

\begin{figure*}[!t]
        \centering
        \includegraphics[width=0.97\linewidth]{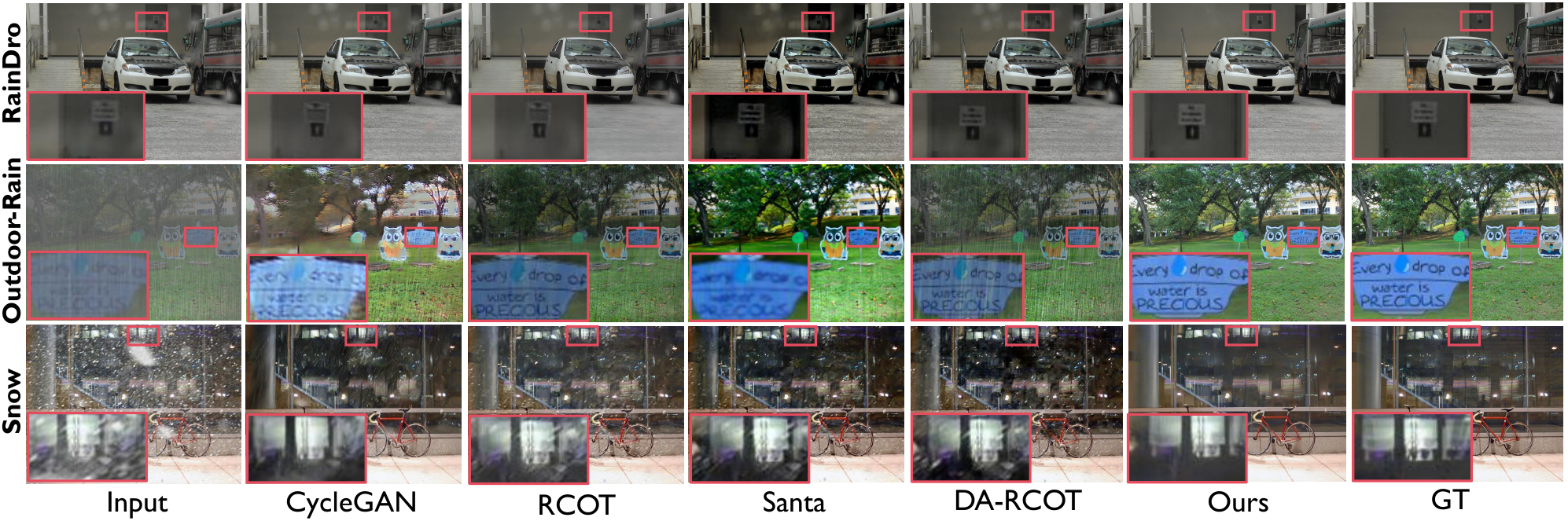}
         
        \caption{Qualitative comparisons on the allweather dataset. Red boxes highlight key visual differences.}
        \label{fig:allweather}
\end{figure*}

In the contrastive setting, the restored image from our WeatherCycle acts as the \textit{anchor}, the clean-domain image as the \textit{positive}, and the degraded input as the \textit{negative}. A DINOV2 \cite{oquab2023dinov2} is used to extract corresponding feature embeddings. We formulate the DACR loss:
\begin{equation}
\mathcal{L}_{\text{dacr}} = \sum_{i \in \mathcal{H}} \left[ \log \frac{\exp \left( \text{sim}(z^i_\text{anchor}, z^i_\text{pos}) / \tau \right)}{\sum\limits_{j} \omega_{j} \cdot \exp  \left( \text{sim}(z^i_\text{anchor}, z^j_\text{neg}) / \tau \right)} \right],
\end{equation}
where $z^i_r$ and $z^i_c$ denote the depth features of the restored and clean images respectively, $\text{sim}(\cdot)$ indicates cosine similarity, $\tau=0.1$ is a temperature scaling factor, and $\mathcal{H}$ denotes the set of selected samples.

To account for the varying difficulty of negative samples, we define a difficulty-aware weighting term $\omega_j$ as:
\begin{equation}
\omega_j = 
\begin{cases}
1 & \text{if } z^j_{\text{neg}} \text{ is an easy negative}, \\
\alpha & \text{if } z^j_{\text{neg}} \text{ is a hard negative}, \\
\beta & \text{if } z^j_{\text{neg}} \text{ is a very hard negative},
\end{cases}
\quad \text{with } \beta > \alpha > 1.
\label{eq:dacr_weights}
\end{equation}
where difficulty-aware weights as $\alpha = 3$ for hard negatives and $\beta = 5$ for very hard negatives, which provide a good balance between contrastive pressure and training stability.

\textbf{Cycle Consistency Loss.} 
To ensure reversible mappings between the degraded and clean domains, we impose pixel-wise and frequency-domain consistency:
\begin{equation}
\begin{aligned}
\mathcal{L}_{\text{cyc}} = &~\| \mathcal{D}_{x}^{d2c2d} - \mathcal{D}_x \|_1 
+ \| \mathcal{C}_{x}^{c2d2c} - \mathcal{C}_x \|_1 \\
&+ \beta\lambda_{\text{fft}} (\mathcal{D}_{x}^{d2c2d}, \mathcal{D}_x) 
+ \beta\lambda_{\text{fft}}(\mathcal{C}_{x}^{c2d2c} ,\mathcal{C}_x),
\end{aligned}
\end{equation}
where $\lambda_{\text{fft}}$ denotes the Fourier loss \cite{sgdn}, and $\beta$ is balances factor with 0.1.

\textbf{Total Loss.} we formulate the total objective as a weighted combination of cycle consistency and depth contrastive regularization losses:
\begin{equation}
\mathcal{L}_{\text{total}} = \lambda_{\text{cyc}} \mathcal{L}_{\text{cyc}} + \lambda_{\text{dacr}} \mathcal{L}_{\text{dacr}},
\end{equation}
where $\lambda_{\text{cyc}}$ and $\lambda_{\text{dacr}}$ are the weight coefficients of each loss term with 1 and 0.8.

\begin{table*}[!t]
    \centering
    \renewcommand{\arraystretch}{1.1}
       \caption{Comparison to state-of-the-art on composited degradations in the CDD11 dataset.}
  
    \setlength{\tabcolsep}{1.7pt}
    \scalebox{0.96}{
    \begin{tabular}{p{0.47cm}|c|cc|cc|cc|cc|cc|cc|cc|cc|cc|cc|cc|cc}
 \Xhline{1.2pt} 
        \multirow{2}{*}{\rotatebox[origin=c]{90}{Type}} & \multirow{2}{*}{Methods} 
          & \multicolumn{8}{c}{\textit{CDD11-Single}} 
          & \multicolumn{10}{c}{\textit{CDD11-Double}}
          & \multicolumn{4}{c}{\textit{CDD11-Triple}} 
          & \multirow{2}{*}{Avg} \\ 
          
         \cmidrule(lr){3-10}  \cmidrule(lr){11-20} \cmidrule(lr){21-24}
         
          ~ & ~ & \multicolumn{2}{c}{Low(L)}   & \multicolumn{2}{c}{Haze(H)}  &\multicolumn{2}{c}{Rain(R)}  & \multicolumn{2}{c}{Snow(S)}  &\multicolumn{2}{c}{L+H}  & \multicolumn{2}{c}{L+R}  & \multicolumn{2}{c}{L+S} & \multicolumn{2}{c}{H+R}  & \multicolumn{2}{c}{H+S}  & \multicolumn{2}{c}{L+H+R}  &  \multicolumn{2}{c}{L+H+S}  \\ \hline

         \multirow{5}{*}{\rotatebox[origin=c]{90}{Paired}} & AirNet  & 
         24.83 &\cc{.778} & 
         24.21 &\cc{.951} & 
         26.55&\cc{.891} & 
         26.79&\cc{.919}
        & 23.23&\cc{.779} & 
        22.82&\cc{.710} & 
        23.29&\cc{.723} & 
        22.21&\cc{.868} & 
        23.29&\cc{.901}
        & 21.80&\cc{.708} & 
        22.24&\cc{.725} & 
        23.75&\cc{.814}\\

        ~ & PromptIR   
        & \oo{\textbf{26.32}}&\oo{\textbf{.805}} & 
        26.10&\cc{.969} & 
        31.56&\cc{.946} & 
        31.53&\cc{.960} & 
        \oo{\textbf{24.49}}&\cc{.789} & 
        25.05&\cc{.771} & 
        24.51&\cc{.761} & 
        24.54&\cc{.924} & 
        23.70&\cc{.925} 
        & 23.74&\cc{.752} & 
        23.33&\cc{.747} & 
        25.90&\cc{.850}\\

        ~ & WGWSNet   & 
        24.39&\cc{.774} & 
        \oo{\textbf{27.90}}&\oo{\textbf{.982}} & 
        \oo{\textbf{33.15}}&\oo{\textbf{.964}} & 
        \oo{\textbf{34.43}}&\oo{\textbf{.973}} 
        & 24.27&\oo{\textbf{.800}} & 
        \oo{\textbf{25.06}}&\oo{\textbf{.772}} & 
        \oo{\textbf{24.60}}&\oo{\textbf{.765}} & 
        \oo{\textbf{27.23}}&\oo{\textbf{.955}} & 
        \oo{\textbf{27.65}}&\oo{\textbf{.960}} & 
        \oo{\textbf{23.90}}&\oo{\textbf{.772}} & 
        \oo{\textbf{23.97}}&\oo{\textbf{.771}} & 
        \oo{\textbf{26.96}}&\oo{\textbf{.863}} \\ 

        ~ & WeatherDiff  & 
        23.58&\cc{.763} & 
        21.99&\cc{.904} & 
        24.85&\cc{.885} & 
        24.80&\cc{.888} & 
        21.83&\cc{.756} & 
        22.69&\cc{.730} & 
        22.12&\cc{.707} & 
        21.25&\cc{.868} & 
        21.99&\cc{.868} & 
        21.23&\cc{.716} & 
        21.04&\cc{.698} & 
        22.49&\cc{.799} \\ 

       ~ & OneRestore  & 
       \rr{\textbf{26.48}} &\rr{\textbf{.826}} & 
       \rr{\textbf{32.52}}&\rr{\textbf{.990}} & 
       \rr{\textbf{33.40}}&\rr{\textbf{.964}} & 
       \rr{\textbf{34.31}}&\rr{\textbf{.973}} & 
       \rr{\textbf{25.79}}&\rr{\textbf{.822}} &
       \rr{\textbf{25.58}}&\rr{\textbf{.799}} & 
       \rr{\textbf{25.19}}&\rr{\textbf{.789}} & 
       \rr{\textbf{29.99}}&\rr{\textbf{.957}} & 
      \rr{\textbf{30.21}}&\rr{\textbf{.964}} & 
       \rr{\textbf{24.78}}&\rr{\textbf{.788}} & 
       \rr{\textbf{24.90}}&\rr{\textbf{.791}} & 
       \rr{\textbf{28.47}}&\rr{\textbf{.878}} \\
        
        
        \hline  \hline 
        
         \multirow{5}{*}{\rotatebox[origin=c]{90}{Unpaired}} & CycleGAN  & 
            21.31 & \cc{.681} & 
            22.92 & \cc{.712} & 
            23.17 & \cc{.715} & 
            25.44 & \cc{.699} & 
            17.83 & \cc{.651} & 
            19.76 & \cc{.668} & 
            21.05 & \cc{.593} & 
            20.14 & \cc{.728} & 
            22.63 & \cc{.719} & 
            19.09 & \cc{.573} & 
            17.27 & \cc{.523} &
            20.97 & \cc{.687} \\ 
         

        

        ~ & Santa   & 
        22.11 & \cc{.718} & 
        25.44 & \cc{.852} & 
        28.96 & \cc{.826} & 
        27.96 & \cc{.814} & 
        19.97  & \cc{.607} & 
        21.34 & \cc{.690} & 
        21.27 & \cc{.630} & 
        22.56 & \cc{.871} & 
        23.44 & \cc{.855} & 
        21.19 & \cc{.653} & 
        19.83 & \cc{.633} & 
        23.01 & \cc{.741}\\

        
         ~ & RCOT  & 
         24.55 & \cc{.760} & 
         26.31 & \cc{.943} & 
         29.66 & \cc{\textcolor{clorange}{.920}} &
         28.81 & \cc{.899} & 
         20.38 & \cc{.675} & 
         22.61 & \cc{.727} & 
         23.19 & \cc{.719} & 
         22.99 & \cc{.863} & 
         22.98 & \cc{.930} & 
         21.51 & \cc{.707} &
         21.49 & \cc{.683} & 
         24.04 & \cc{.803} \\
         
          ~ & DA-RCOT  & 
           \textcolor{clorange}{25.53} & \cc{\textcolor{clorange}{.763}} &
          \textcolor{clorange}{26.60} & \cc{\textcolor{clorange}{.945}} & 
          \textcolor{clorange}{31.05} & \cc{\textcolor{red!50}{\textbf{.925}}} &
          \textcolor{clorange}{30.72} & \cc{\textcolor{clorange}{.933}} & 
          \textcolor{red!50}{\textbf{21.99}} & \cc{\textcolor{clorange}{.709}} &
          \textcolor{red!50}{\textbf{23.13}} & \cc{\textcolor{red!50}{\textbf{.770}}}& 
          \textcolor{clorange}{24.39} & \cc{\textcolor{clorange}{.727}} & 
          \textcolor{clorange}{23.08}& \cc{\textcolor{clorange}{.875}} & 
          \textcolor{clorange}{24.66} & \cc{\textcolor{clorange}{.935}} & 
          \textcolor{clorange}{22.02} & \cc{\textcolor{clorange}{.709}} & 
          \textcolor{clorange}{21.97} & \cc{\textcolor{clorange}{.698}} & 
          \textcolor{clorange}{25.01} & \cc{\textcolor{clorange}{.817}} \\ \cline{2-26} 
        

          ~ & Ours  
          & \textcolor{red!50}{\textbf{25.75}} & \cc{\textcolor{red!50}{\textbf{.766}}} & 
          \textcolor{red!50}{\textbf{26.85}} & \cc{\textcolor{red!50}{\textbf{.947}}} &
          \textcolor{red!50}{\textbf{31.40}} & \cc{\textcolor{red!50}{\textbf{.925}}} & 
          \textcolor{red!50}{\textbf{32.16}} &\cc{\textcolor{red!50}{\textbf{.937}}} & 
          \textcolor{clorange}{21.96}& \cc{\textcolor{red!50}{\textbf{.711}}} & 
          \textcolor{clorange}{22.82} & \cc{\textcolor{clorange}{.769}} &  
          \textcolor{red!50}{\textbf{24.55}} & \cc{\textcolor{red!50}{\textbf{.737}}} & 
          \textcolor{red!50}{\textbf{25.63}} & \cc{\textcolor{red!50}{\textbf{.927}}} & 
          \textcolor{red!50}{\textbf{25.33}} & \cc{\textcolor{red!50}{\textbf{.937}}} & 
          \textcolor{red!50}{\textbf{22.92}} & \cc{\textcolor{red!50}{\textbf{.718}}} & 
          \textcolor{red!50}{\textbf{22.53}} & \cc{\textcolor{red!50}{\textbf{.705}}} & 
          \textcolor{red!50}{\textbf{25.63}} & \cc{\textcolor{red!50}{\textbf{.825}}}\\  \Xhline{1.2pt} 
    \end{tabular}
  } 
    \label{table:cdd}
\end{table*}

\begin{figure*}[!t]
        \centering
        \includegraphics[width=0.97\linewidth]{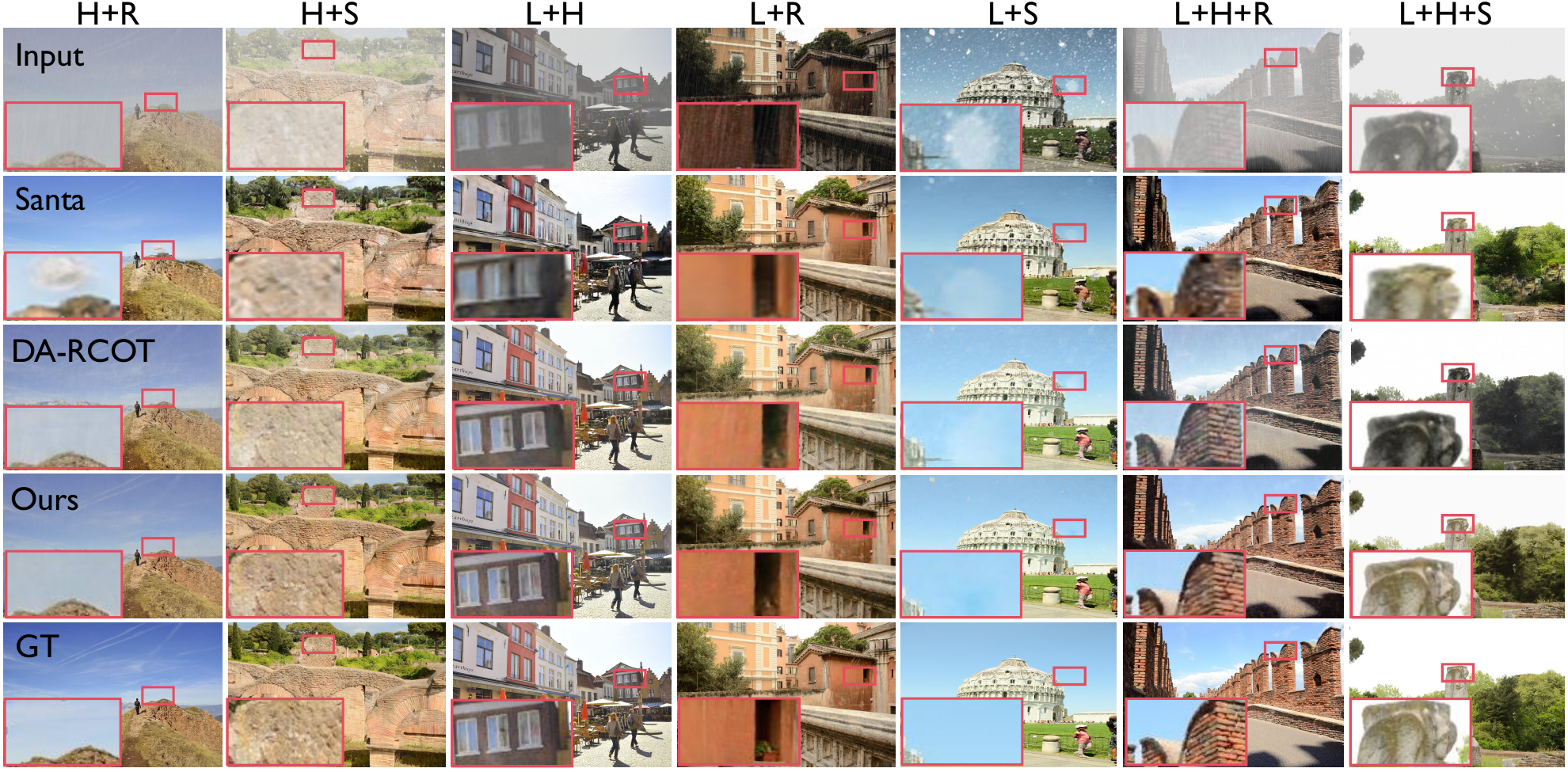}
        \vskip 0in
        \caption{Qualitative comparisons on the CDD11 dataset under various compound degradation types. Our method restores clearer structure, finer texture, and more faithful colors. Red boxes highlight key visual differences.}
        \label{fig:CDD11}
\end{figure*}

\section{Experiments}

\subsection{Implementation Details}
All models are implemented in PyTorch and trained using the Adam optimizer with $\beta_1{=}0.9$, $\beta_2{=}0.999$. The initial learning rate is set to $2\times10^{-4}$ and decayed following a cosine annealing schedule without restarts. For all experiments, we train the model for 500K iterations with a batch size of 8. The image resolution is fixed at $256{\times}256$. All training and evaluation are conducted on a 8 NVIDIA RTX 3090 GPUs with 24GB memory. We apply standard data augmentations including random horizontal flipping, random rotation, and color jittering (brightness, contrast, saturation). For unpaired settings, we ensure that input and target domains are randomly shuffled per epoch to enhance generalization.

\subsection{Experiment Settings}
To comprehensively evaluate our model, we conduct experiments on two mutli-weather restoration datasets: as in prior works \cite{allinone, t3diffweather}, we use the AllWeather dataset, which comprises 18,069 images drawn from three widely-used benchmarks: Snow100K \cite{liu2018desnownet}, Outdoor-Rain \cite{li2019heavy}, and Raindrop \cite{qian2018attentive}. The Composite Degradation Dataset (CDD11) dataset \cite{guo2025onerestore} consists of 15,190 images in total, including 12,990 for training and 2,200 for testing. It covers 11 degradation settings, spanning single (L, H, R, S), dual (e.g., L+R, H+S), and triple (L+H+R, L+H+S) weather combinations. All performances are assessed using standard PSNR and SSIM metrics.

\begin{figure*}[!t]
    \centering 
    \includegraphics[width=0.97\linewidth]{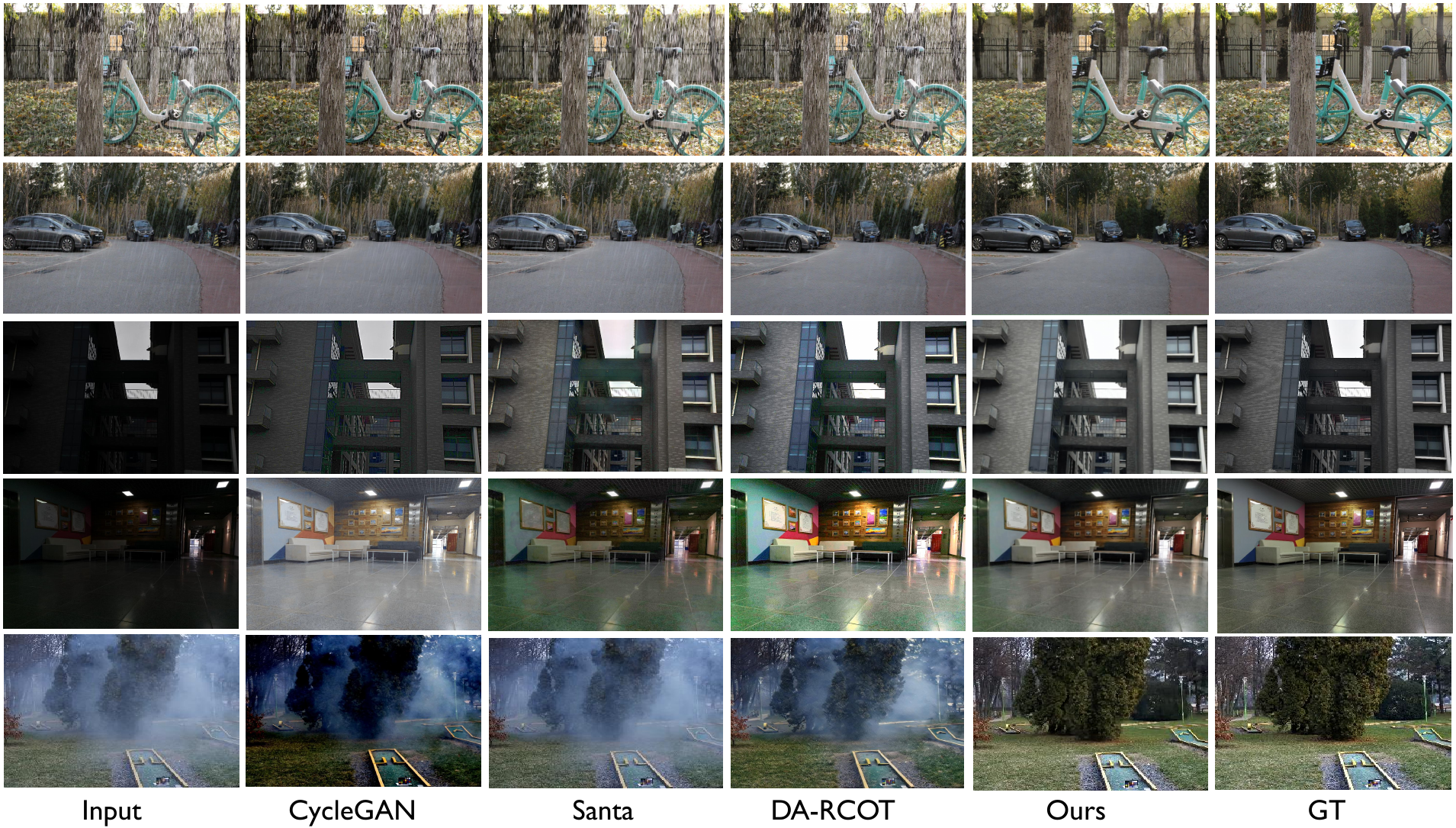}
    \caption{Real-world generalization results on naturally degraded samples. Our model restores sharp and natural-looking results despite unseen degradations.} 
    \label{figure:real_results}
\end{figure*}


\subsection{Quantitative comparison.}
We compare our unpaired restoration framework with both paired supervision methods and existing unpaired baselines. All models are trained strictly following their official implementations and protocols to ensure fair comparison.

Table~\ref{table:all_weather_result} reports the performance of all competing methods on the AllWeather dataset. Compared to WGWS-Net~\cite{zhu2023learning}, our method achieves a notable improvement of 3.15 dB on Outdoor-Rain. Moreover, it surpasses the recent strong supervised baseline TransWeather-TUR~\cite{wu2025debiased} by 0.39 dB on Snow100K-S. In particular, our model ranks first among all unpaired methods on four sub-datasets and yields the highest overall average PSNR and SSIM, underscoring the effectiveness of our luminance-guided unpaired learning strategy. As shown in Table~\ref{table:cdd}, our method significantly outperforms both classical unsupervised models (e.g., CycleGAN \cite{cyclegan}, CUT \cite{cut}) and recent all-in-one approaches (e.g., Santa \cite{santa}, DA-RCOT \cite{darcot}) on the CDD11 benchmark, which includes single, double, and triple weather degradations. Notably, our method achieves the best average PSNR and SSIM scores among all unpaired methods, especially under more complex dual and triple degradation settings. In addition, WeatherCycle also significantly outperforms the recent unified model WeatherDiff \cite{weatherdiff} in 11 settings.

\begin{table}[t]
    \normalsize
      \caption{Quantitative dehazing performance comparisons of different methods. Red and blue colors indicate the 1st and 2nd ranks among the unsupervised methods, respectively.}
      \label{table.Unsuperviseddehazing}
      \centering
      \setlength{\tabcolsep}{3pt}
      \scalebox{0.76}{
      \begin{tabular}{p{0.3cm}|c|c|cccc|cc}
        \toprule
        \multirow{2}{*}{\rotatebox[origin=c]{90}{Type}}  & \multirow{2}{*}{Methods} & \multirow{2}{*}{Venue}  & \multicolumn{2}{c}{SOTS-Indoor } & \multicolumn{2}{c|}{SOTS-Outdoor} & \multicolumn{2}{c}{NH-Haze}   \\
           
        ~ & ~ & ~ & PSNR & SSIM & PSNR & SSIM & PSNR & SSIM    \\
         \cmidrule(r){1-9}
        \multirow{3}{*}{\rotatebox[origin=c]{90}{Paired}} & FFANet \cite{ffanet} & ECCV'20 & 36.39 & 0.988 & 33.57& 0.984 & 20.00 & 0.822  \\
    
        
        & C$^2$PNet \cite{c2pnet} & CVPR'23  & 42.56 & 0.995 & 36.68 & 0.990 & 21.19 & 0.833  \\
        
        & SGDN \cite{sgdn} & AAAI'25  & 42.19 & 0.998 & 36.22 & 0.986 & 21.71 & 0.834  \\ \midrule
        
        \multirow{8}{*}{\rotatebox[origin=c]{90}{Unpaired}} & DCP \cite{dcp} & CVPR'09  & 13.10 & 0.699 & 19.13 & 0.815 & 14.90 & 0.668  \\
        
        & CycleGAN \cite{cyclegan} & ICCV'17  & 21.34 & 0.898 & 20.55 & 0.856 & 13.95 & 0.689  \\

         & CUT \cite{cut} & ECCV'20  & 24.30 & 0.911 & 23.67 & 0.904 & 15.92 & 0.758 \\
        
        & YOLY \cite{yoly} & IJCV'21  & 15.84 & 0.819 & 14.75 & 0.857 & 13.38 & 0.595  \\


        & D$^4$ \cite{d4} & CVPR'22  & 25.42 & 0.932 & 25.83 & 0.956 & 14.52 & 0.709 \\
        
        & NSDNet \cite{nsdnet} & TCSVT'25  & 26.34 & 0.937 & 25.77 & 0.951 & 17.36 & 0.761 \\
       
        &  ODCR \cite{odcr} & CVPR'24  & 26.32 & \underline{0.945} & 26.16 &\underline{0.960}& \underline{17.56} & \underline{0.766}  \\

         &  FrDiff \cite{liu2025frequency} & ICCV'25  &\underline{27.43} & \textbf{0.957} & \underline{22.75} &0.914& - & -  \\
    
         \cmidrule{1-9}
         
         &  Ours & - & \textbf{27.64} & 0.940 & \textbf{28.03} & \textbf{0.965} & \textbf{18.71} & \textbf{0.770} \\
        \bottomrule
      \end{tabular}
      }  
\end{table}

\begin{table}[t]
    \normalsize
      \caption{Quantitative deraining performance comparisons of different methods. Red and blue colors indicate the 1st and 2nd ranks among the unsupervised methods, respectively.}
      \label{table.Unsupervisedderainingresults}
      \centering
      \setlength{\tabcolsep}{1.54pt}
      \scalebox{0.78}{
      \begin{tabular}{p{0.3cm}|c|c|cccc|cc}
        \toprule
        \multirow{2}{*}{\rotatebox[origin=c]{90}{Type}}& \multirow{2}{*}{Methods} & \multirow{2}{*}{Venue}  & \multicolumn{2}{c}{Rain100L } & \multicolumn{2}{c|}{Rain100H } & \multicolumn{2}{c}{RealRS}   \\
          
        ~ & ~ & ~ & PSNR & SSIM & PSNR & SSIM & PSNR & SSIM    \\
  
        \midrule
        \multirow{3}{*}{\rotatebox[origin=c]{90}{Paired}}  & MPRNet   \cite{mprnet} & CVPR'21 &34.95&0.959&28.53&0.872&23.24&0.656\\    
         
         ~ &  Restormer \cite{restormer} & CVPR'22 &37.57&0.974&29.46&0.889&   22.96 & 0.648  \\
          
         ~ &  PromptIR \cite{promptir} & NIPS'23 &38.34&0.983&28.69&0.877&  22.48& 0.625 \\
         
     
        \midrule
        
      \multirow{6}{*}{\rotatebox[origin=c]{90}{Unpaired}} & CycleGAN \cite{cyclegan} &ICCV'17 & 24.61 & 0.834 &20.59&0.704& -& - \\   
      
       ~ & DerainCycleGAN \cite{deraincyclegan} &TIP'21 &  31.49 &  0.936 &  21.13 & 0.710 & 18.75 & 0.460  \\   
       
       ~ &NLCL  \cite{nlcl} &CVPR'22 & 20.50 & 0.719 &22.31&0.728&17.00 &0.477\\
       
        ~ &DCD-GAN \cite{dcdgan} & CVPR'22 &  31.82 & 0.941 & 22.47 & 0.753 & 22.13 &0.610  \\
        
        ~ & CSUD \cite{csud} & CVPR'25 & \underline{33.28} & \underline{0.954} & \underline{24.42} & \underline{0.808} & \underline{22.54} & \underline{0.613 } \\
        \cmidrule(r){1-9}
        ~ & Ours  & -   & \textbf{34.41} & \textbf{0.960} & \textbf{25.72} & \textbf{0.826} & \textbf{24.03} & \textbf{0.670}  \\
        \bottomrule
      \end{tabular}
      }  
\end{table}

\begin{table}[t]
    \normalsize
      \caption{Quantitative comparisons on the paired LOL and LSRW datasets.}
      \setlength{\tabcolsep}{2pt}
      \label{table_low}
      \centering
      \scalebox{0.76}{
      \begin{tabular}{p{0.3cm}|c|c|cccc|cc}
        \toprule
        \multirow{2}{*}{\rotatebox[origin=c]{90}{Type}}& \multirow{2}{*}{Methods} & \multirow{2}{*}{Venue}  & \multicolumn{2}{c}{LOL} & \multicolumn{2}{c|}{LSRW} & \multicolumn{2}{c}{DICM}   \\
          
        ~ & ~ & ~ & PSNR & SSIM & PSNR & SSIM & NIQE $\downarrow$  & PI $\downarrow$    \\  \midrule
        
        \multirow{2}{*}{\rotatebox[origin=c]{90}{Paired}}  & SMG  \cite{smg} &CVPR'23  & 23.81 & 0.809 &17.57&0.538& 6.224 &4.228\\    
          
         ~ &  GSAD \cite{gsad}&NIPS'23 &22.02&0.848&17.41&0.507&  4.496& 3.593 \\
         

        \midrule
          \multirow{10}{*}{\rotatebox[origin=c]{90}{Unpaired}} & Zero-DCE \cite{zerodce} &CVPR'20& 14.86 & 0.562 &15.86 &0.443& 3.951& 3.149 \\   
      
       ~ & EnlightenGAN  \cite{enlightengan}& TIP'21&  17.60 &  0.653 &  17.10 & 0.463 & 3.832 & 3.256 \\   
       
       ~ & RUAS  \cite{ruas} &CVPR'21 & 16.40 & 0.503 & 14.27 & 0.461 & 7.306 & 5.700 \\
       
        ~ & SCI \cite{sci} &CVPR'22&  14.78 & 0.525 & 15.24 & 0.419 & 4.519 & 3.700  \\
        
        ~ & GDP \cite{gdp}  &CVPR'23& 15.89 & 0.542 & 12.88 & 0.362 & 4.358 & 3.552  \\

        ~ & PairLIE  \cite{pairlie} &CVPR'23& 19.51 & 0.731 & 17.60 & 0.501 & 4.282 & 3.469  \\

         ~ & NeRco \cite{nerco}  &ICCV'23& 17.73 & 0.740 & 17.84 & 0.535 & 4.107 & 3.345  \\

         ~ & LightenDiffusion \cite{lightendiffusion}  &ECCV'24& 20.45 & 0.803 & 18.55 & \underline{0.539} & \textbf{3.724} & \underline{3.144}   \\

          ~ & DPLUT \cite{lin2025dplut}  &AAAI'25 & \underline{20.66} & 0.740 & \underline{18.91} & 0.530 & - & -   \\

           ~ & AGLLdiff \cite{lin2025aglldiff}  & AAAI'25 & 19.83 & \underline{0.806}& 17.35 & \underline{0.544} & 4.099 & 3.561   \\
           
        \cmidrule(r){1-9}
        ~ & Ours  &- & \textbf{21.09} & \textbf{0.819} & \textbf{19.39} & \textbf{0.554} & \underline{3.799} & \textbf{3.046} \\
        \bottomrule
      \end{tabular}
      }  
\end{table}

\subsection{Cross-Dataset Generalization}

To demonstrate the generalization capability of our method beyond the All-Weather benchmark, we conduct experiments on multiple widely-used datasets covering haze, rain, and low-light degradations. These datasets include SOTS \cite{reside}, NH-Haze \cite{nhhaze} for dehazing, Rain100L/H and RainDSRealRS \cite{rain100} for deraining, and LOL \cite{lol}, LSRW \cite{lsrw}, and DICM \cite{dicm} for low-light enhancement. We compare our model with both paired and unpaired state-of-the-art methods. The results are summarized in Tables~\ref{table.Unsuperviseddehazing}–\ref{table_low}.

\noindent\textbf{Dehazing Performance.}  
Table~\ref{table.Unsuperviseddehazing} shows the quantitative dehazing results. Our method achieves the best overall performance across all test sets among unpaired approaches, outperforming ODCR \cite{odcr}, NSDNet \cite{nsdnet}, and CUT \cite{cut}. Notably, our model even surpasses some paired methods on SOTS-Outdoor. On the real-world NH-Haze dataset, our method achieves 18.71 PSNR and 0.770 SSIM, showing strong robustness under complex real hazy conditions.

\noindent\textbf{Deraining Performance.}  
In Table~\ref{table.Unsupervisedderainingresults}, we report deraining results on both synthetic (Rain100L/H) and real-world (RainDSRealRS) datasets. Our method consistently achieves the highest scores among unpaired approaches. Particularly, it outperforms CSUD \cite{csud} by +0.75 dB on Rain100H and +1.49 dB on RainDSRealRS, indicating its capability in handling both light and heavy rain scenarios. Compared to DCD-GAN and DerainCycleGAN, our approach yields more stable and higher quality restoration across diverse rain types.

\noindent\textbf{Low-Light Enhancement.}  
We also evaluate our method on paired LOL and LSRW datasets as well as the unpaired DICM benchmark (Table~\ref{table_low}). Despite being trained without direct supervision, our model outperforms existing unpaired methods such as LightenDiffusion \cite{lightendiffusion}, NeRco \cite{nerco}, and Zero-DCE \cite{zerodce} on both PSNR and SSIM. It also yields the best perceptual quality on the DICM dataset in terms of NIQE (3.799) and PI (3.046), confirming its ability to produce visually pleasing results under real low-light conditions.

\begin{figure*}[!t]
    \centering 
    \includegraphics[width=0.87\linewidth]{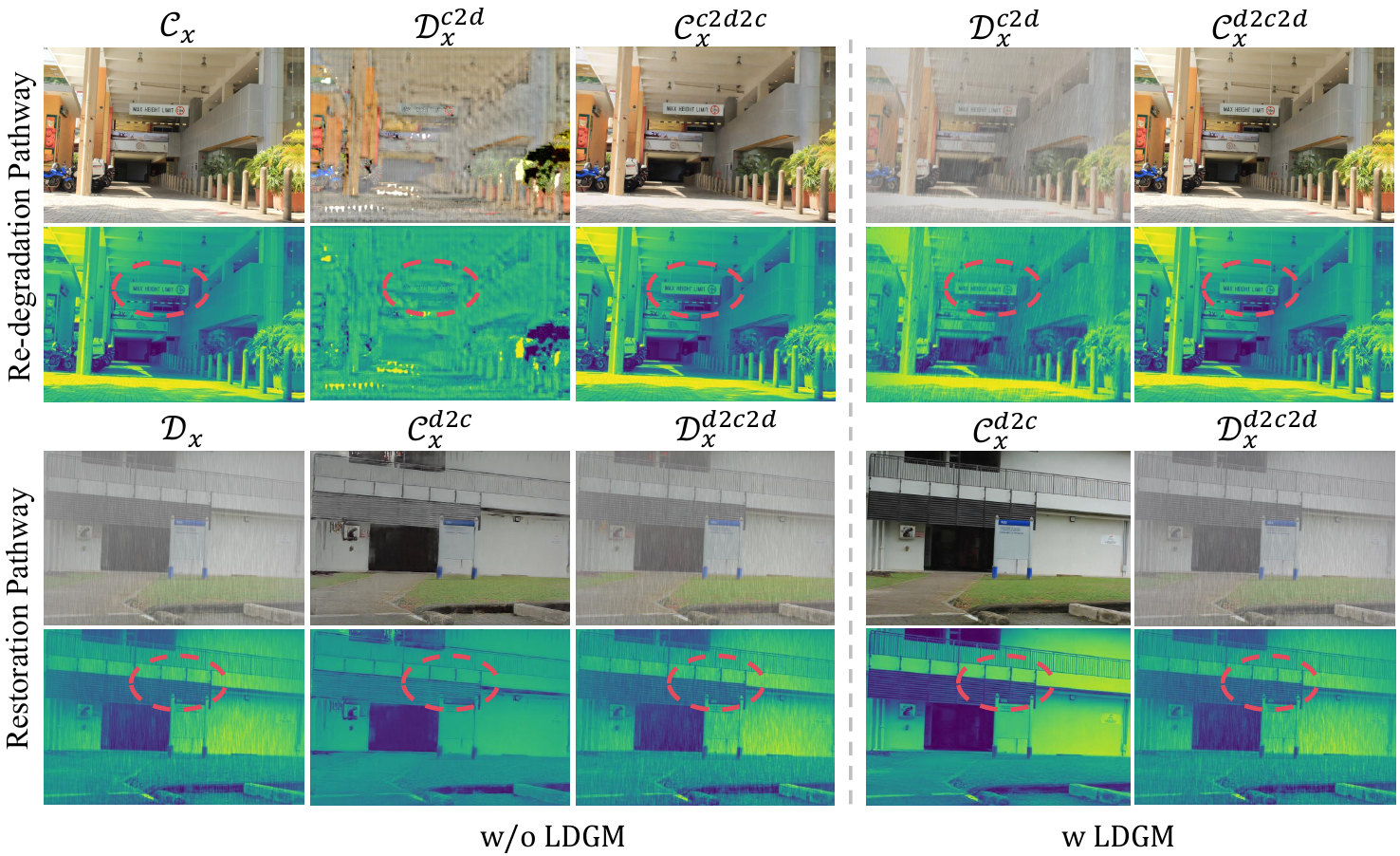}
    \caption{Visual comparison of WeatherCycle with and without the proposed LDGM in both re-degradation (top) and restoration (bottom) pathways. $\mathcal{C}_x$, $\mathcal{D}_x$ denote clean and degraded images. $\mathcal{D}_{c2d}$ and $\mathcal{C}_{d2c}$ are translated outputs, while $\mathcal{C}_{c2d2c}$, $\mathcal{D}_{d2c2d}$ denote cycle-reconstructed results. The corresponding Y-channel visualizations (below each image) highlight degradation transfer and structural consistency. The LDGM significantly improves content fidelity and degradation realism, particularly in regions with ambiguous lighting or weak texture (circled in red).} 
    \label{figure:appendix_LDGM}
\end{figure*}

\subsection{Visual Comparison.}
Figure~\ref{fig:allweather} presents qualitative comparisons of advanced unsupervised methods on the All-Weather dataset. Our approach consistently produces clearer and more visually coherent results across various degradation types, including raindrops, snow, and rain-fog. While DA-RCOT \cite{darcot} fails to fully eliminate degradation in certain regions, methods like CycleGAN \cite{cyclegan} and Santa \cite{santa} yield overly blurred outputs that miss fine background details. In contrast, our model preserves smooth transitions and accurate object structures. Figure~\ref{fig:CDD11} shows qualitative results on the CDD11 dataset under various compound degradation types. Compared to other unsupervised methods, Santa and DA-RCOT often leave residual artifacts or oversmooth details, while our method delivers sharper edges and clearer scene content, especially in challenging L+H+R, L+H+S conditions.

In Figure~\ref{figure:real_results}, we evaluate our model on real-world degraded images from uncontrolled settings. These scenes include challenging lighting, weather overlays, and unknown camera effects. Despite lacking paired supervision or explicit real degradation modeling, our model generalizes well and produces visually pleasing results with restored contrast, reduced noise, and realistic colors. The performance gap between baselines and our approach further highlights the strength of our unpaired framework in practice.

\begin{table}[!t]
    \normalsize
      \caption{Ablation studies on the allweather dataset.}
      \vspace{-0.1cm}
      \label{table:ablation}
      \centering
      \scalebox{0.75}{
      \begin{tabular}{l|cccccc}
        \toprule
         \multirow{2}{*}{Methods}   & \multicolumn{2}{c}{Snow} & \multicolumn{2}{c}{Raindrop} & \multicolumn{2}{c}{Outdoor-Rain}   \\
           
         ~  & PSNR & SSIM & PSNR & SSIM & PSNR & SSIM    \\
         \cmidrule(r){1-7}

         w/o $\mathcal{G}_{d2c}$  & 26.61 & 0.863 & 24.72 & 0.837 & 24.39 & 0.873 \\
        
         w/o $\mathcal{G}_{c2d}$ & 28.05 & 0.883 & 26.84 & 0.861 & 25.73 & 0.902 \\
         
        w/o $\mathcal{J}_{c2d}$ & 30.64 & 0.907 & 28.26 & 0.886 & 27.38 & 0.909\\
        
         w/o LDGM & 30.33 & 0.909 & 28.71 & 0.893 & 27.89 & 0.911 \\
       
          w/o $\mathcal{L}_{\text{dacr}}$ & 31.41 & 0.912 & 29.36 & 0.895 & 28.22 & 0.918  \\
    
         \cmidrule{1-7}
         
          Ours  & \textbf{31.63} & \textbf{0.919} & \textbf{29.56} & \textbf{0.900} & \textbf{28.46} & \textbf{0.929} \\
        \bottomrule
      \end{tabular}
      }  
\end{table}

\begin{figure}[!t]
        \centering
        \includegraphics[width=1\linewidth]{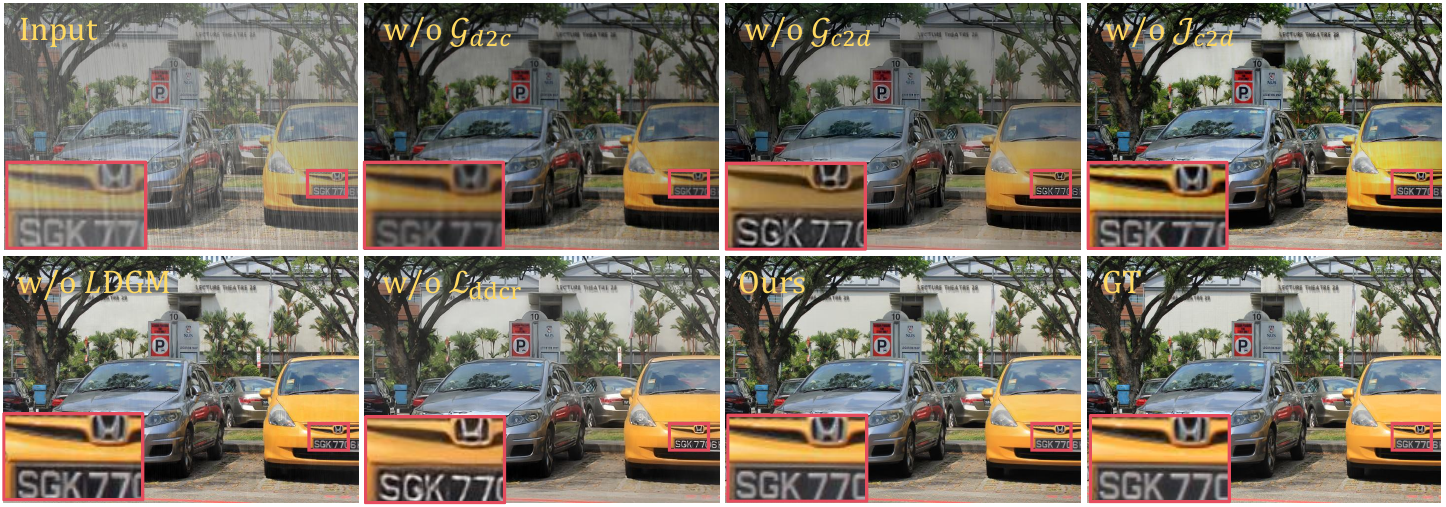}
        \vspace{-0.6cm}
        \caption{Ablation of each modules in our WeatherCycle.}
        \label{fig:ablation_psnr}
\end{figure}


\begin{figure}[!t]
    \centering 
    \includegraphics[width=1\linewidth]{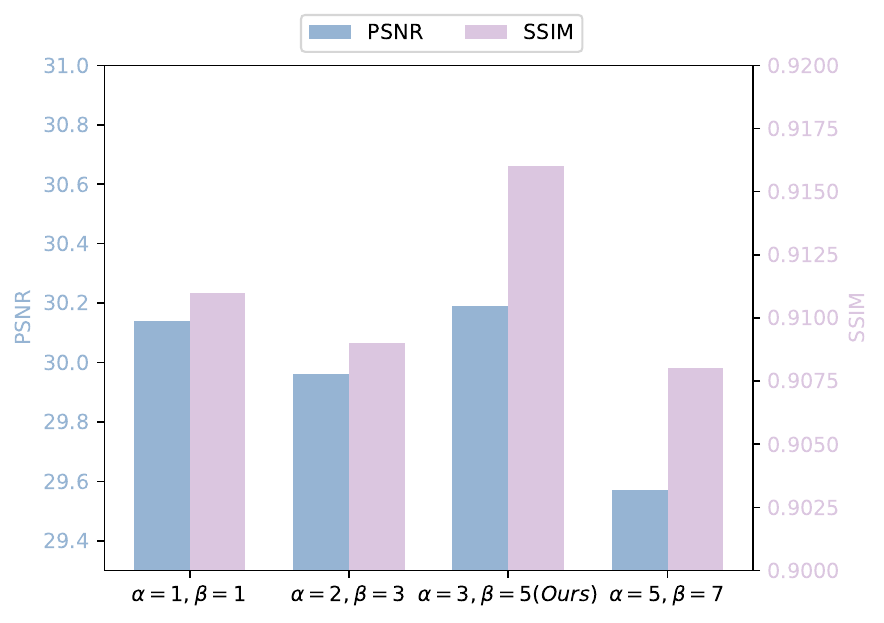}
    \caption{Ablation study on difficulty-aware weighting in DACR. The combination $\alpha{=}3, \beta{=}5$ (Ours) achieves the best performance in both PSNR and SSIM.} 
    \label{fig:dacr_ablation}
\end{figure}

           
         
        
       
    
         

        

\section{Ablation Studies}
In this section, we conduct comprehensive ablation studies to evaluate the contribution of each components of our WeatherCycle on the allweather dataset.

\subsection{Effectiveness of Proposed Modules.}
As summarized in Table~\ref{table:ablation}. Removing $\mathcal{G}_{d2c}$ replaces it with a vanilla UNet significantly degrades performance, leads to a notable PSNR reduction of 4.07~dB on Outdoor-Rain, confirming the importance of separating illumination and color information. Without $\mathcal{G}_{c2d}$, the backward clean-to-degraded mapping lacks domain consistency, resulting in a 2.72~dB drop on Raindrop. Eliminating the auxiliary degradation injection $\mathcal{J}_{c2d}$ weakens the bidirectional guidance and leads to a 1.08~dB decrease on Snow. Removing LDGM causes a 0.57~dB decline on Outdoor-Rain, indicating the importance of recurrent degradation prior modeling. Finally, removing the contrastive loss $\mathcal{L}_{\text{dacr}}$ results in a 2.17\% SSIM drop on Outdoor-Rain, highlighting its role in enhancing structure-aware representation learning. These results collectively demonstrate that all components contribute positively to the final performance. Figure~\ref{fig:ablation_psnr} qualitatively verifies the role of each component in \textit{WeatherCycle}. Removing the restoration path $\mathcal{G}_{d2c}$ leaves strong haze and low contrast—the license plate characters and grille edges remain indistinct. Without the re-degradation path $\mathcal{G}_{c2d}$, the cycle constraint weakens; the result looks over-smoothed with noticeable color shift on the hood and windshield, revealing an ill-posed clean degraded mapping. Dropping the joint constraint $\mathcal{J}_{c2d}$ introduces ringing/ghosting near high-frequency structures and unstable tones around specular regions. In contrast, the full model recovers the sharpest plate characters, natural colors, and consistent local textures, matching the GT most closely—consistent with the quantitative trends in our ablations.


\subsection{Effectiveness of LDGM in WeatherCycle.}
To better understand the effect of our Random Degradation Guidance Module (LDGM), we present a comparison of restoration and re-degradation results with and without LDGM in Figure \ref{figure:appendix_LDGM}. On the top half (Re-degradation Pathway), a clean image $\mathcal{C}_x$ is translated into a degraded variant $\mathcal{D}_{c2d}$, then cycle-translated back to $\mathcal{C}_{c2d2c}$. The Y-channel visualizations reveal that, without LDGM, the synthesized degradation lacks realistic texture patterns and introduces structural artifacts (e.g., distortion near shadows and signs). In contrast, with LDGM, the degraded outputs better mimic real degradations and preserve spatial layout, enabling more faithful round-trip translation. In the bottom half (Restoration Pathway), a degraded input $\mathcal{D}_x$ is mapped to a clean counterpart $\mathcal{C}_{d2c}$, then degraded back to $\mathcal{D}_{d2c2d}$. Again, LDGM leads to cleaner restoration and better cycle consistency. Notably, in the red-circled regions, LDGM helps preserve wall textures and eliminate illumination inconsistencies.

\begin{table}[!t]
\normalsize
    \caption{Ablation study on the All-Weather dataset across key modules.}
    \vspace{-0.3cm}
    \setlength{\tabcolsep}{2pt}
    \label{table:more_ablation_fixed}
    \centering
    \scalebox{0.76}{
    \begin{tabular}{l|cccccc}
    \toprule
    \multirow{2}{*}{Variant} & \multicolumn{2}{c}{Snow} & \multicolumn{2}{c}{Raindrop} & \multicolumn{2}{c}{Outdoor-Rain} \\
    ~ & PSNR & SSIM & PSNR & SSIM & PSNR & SSIM \\
    \midrule
    RGB decomposition & 30.92 & 0.910 & 28.91 & 0.892 & 27.40 & 0.911 \\
    HSV decomposition & 31.05 & 0.911 & 29.07 & 0.894 & 27.63 & 0.915 \\
    
    \midrule
    Vanilla contrastive loss & 31.23 & 0.914 & 29.28 & 0.896 & 27.96 & 0.921 \\
    
    \midrule
    ResNet50 features & 30.96 & 0.912 & 29.01 & 0.892 & 27.62 & 0.918 \\
    CLIP features & 31.30 & 0.916 & 29.26 & 0.896 & 28.03 & 0.923 \\
    
    \midrule
    Unified restoration & 31.15 & 0.913 & 29.19 & 0.893 & 27.75 & 0.920 \\
    \rowcolor{gray!10} Ours & \textbf{31.63} & \textbf{0.919} & \textbf{29.56} & \textbf{0.900} & \textbf{28.46} & \textbf{0.929} \\
    \bottomrule
\end{tabular}
}
\end{table}

\subsection{Ablations on Color Space, DACR, Semantic Backbone, and Decoupling}
To further validate the effectiveness of each component in our proposed framework, we conduct comprehensive ablation experiments on the All-Weather dataset. Specifically, we analyze four critical factors that influence multi-weather restoration: (1) color space decomposition strategies, (2) difficulty-aware contrastive regularization (DACR), (3) the choice of semantic feature extractor, and (4) the impact of luminance-chrominance decoupling.

Table~\ref{table:more_ablation_fixed} reports the results across three representative degradation types: Snow, Raindrop, and Outdoor-Rain. The results consistently demonstrate that our full model configuration delivers the best overall performance.

\noindent\textbf{Color Space Decomposition.}  
We compare three color spaces—RGB, HSV, and YCbCr—by replacing the decomposition module in our pipeline. The YCbCr decomposition clearly outperforms the other options across all subsets, especially under the Outdoor-Rain setting (28.46 dB vs. 27.40/27.63 dB). This confirms our hypothesis that explicitly separating luminance (Y) and chrominance (Cb/Cr) channels aligns better with real-world weather degradations, which predominantly affect brightness and contrast rather than chroma.

\noindent\textbf{Effectiveness of DACR.}  
We replace DACR with a vanilla contrastive loss (fixed temperature, no classifier). The performance drops across all subsets, particularly in Snow (–0.40 dB) and Outdoor-Rain (–0.50 dB), indicating that DACR not only improves overall recovery but also enhances robustness under challenging, semantically inconsistent degradations. This aligns with our goal to mine and emphasize harder-to-restore regions during training.

\noindent\textbf{Semantic Backbone.}  
We substitute our default DINOv2 backbone with CLIP and ResNet50 to assess the role of semantic priors. DINOv2 achieves the best restoration performance on all three subsets due to its strong structure-awareness and fine-grained visual tokens. CLIP performs reasonably in simpler degradations (e.g., Raindrop), while ResNet50 shows limitations in capturing detailed scene-level semantics, resulting in up to 1 dB performance loss in Outdoor-Rain.

\noindent\textbf{Luminance-Chrominance Decoupling.}  
We compare our decoupled dual-branch architecture (separately modeling Y and Cb/Cr) with a unified restoration model that jointly processes all three channels. As shown in Table~\ref{table:more_ablation_fixed}, the decoupled variant yields consistent improvements (e.g., +0.71 dB in Outdoor-Rain), demonstrating its superiority in handling complex luminance degradation. The unified version tends to blur structural edges and produces inconsistent tone transitions under severe weather.

\subsection{Difficulty-Aware Weighting.}

To investigate the impact of the hyperparameters $\alpha$ and $\beta$ in our difficulty-aware weighting scheme, we conduct an ablation study by varying their values while keeping all other settings fixed. The results are illustrated in Figure~\ref{fig:dacr_ablation}, where we report the performance on a validation split of the All-Weather dataset. As shown in the figure, using equal weights for all negative samples ($\alpha{=}1, \beta{=}1$) yields suboptimal performance, with a PSNR of 30.14 and SSIM of 0.911. This indicates that treating all negatives equally fails to emphasize harder samples that contribute more to representation learning. Gradually increasing the contrastive pressure on harder negatives ($\alpha{=}2, \beta{=}3$) leads to slightly improved alignment, but the full benefit is observed at $\alpha{=}3, \beta{=}5$, where our model achieves the best balance of discrimination and stability, reaching the highest PSNR (30.19) and SSIM (0.916). This validates our hypothesis that scaling the contribution of difficult negatives helps the model focus on ambiguous or confusing cases that are critical for restoration.

\section{Conclusion}
In this paper, we propose an unpaired multi-weather restoration framework based on lumina-chroma decomposition and bidirectional domain translation. By explicitly separating the luminance and chrominance components, our method better preserves structural and color fidelity under adverse weather conditions. The introduced dual generators $\mathcal{G}_{d2c}$ and $\mathcal{G}_{c2d}$, along with the Limina Degradation Guidance Module and Difficulty-Aware Contrastive Regularization, collectively enhance domain consistency and content preservation. Extensive experiments on synthetic and real-world benchmarks demonstrate that our approach outperforms state-of-the-art unsupervised methods across dehazing, deraining, and low-light enhancement tasks.

\ifCLASSOPTIONcaptionsoff
  \newpage
\fi

\bibliographystyle{IEEEtran}
\bibliography{main}

\end{document}